\documentclass[10pt,twocolumn]{article}
\usepackage[margin=0.75in]{geometry}
\usepackage[utf8]{inputenc}
\usepackage[T1]{fontenc}
\usepackage{hyperref}
\usepackage{url}
\usepackage{booktabs}
\usepackage{amsfonts}
\usepackage{amsmath}
\usepackage{graphicx}
\usepackage{xcolor}
\usepackage{multirow}
\usepackage{algorithm}
\usepackage{algorithmic}
\usepackage{subcaption}
\usepackage{natbib}
\usepackage{cleveref}
\usepackage{enumitem}

\graphicspath{{./figures/}}

\newcommand{\method}{EFF}
\newcommand{\fullmethod}{Edit Fidelity Field}

\title{Edit Fidelity Field: Semantics-Aware Region Isolation\\for Training-Free Scene Text Editing}

\author{Guandong Li\\
iFLYTEK\\
\quad leeguandon@gmail.com
\and
Mengxia Ye\\
Aegon THTF}

\begin{document}

\maketitle

% ============================================================================
% ABSTRACT
% ============================================================================
\begin{abstract}
Scene text editing (STE) has achieved remarkable progress in accurately rendering target text through diffusion-based methods. However, we identify a critical yet overlooked problem: \textbf{edit spillover} --- when editing a target text region, existing methods inadvertently modify non-target regions, particularly neighboring text. Through systematic evaluation on 50 real-world scenes across four categories, we reveal that state-of-the-art diffusion editing models exhibit a spillover rate of \textbf{94\%}, meaning nearly all non-target text regions are altered during editing. To address this, we propose the \textbf{\fullmethod{} (\method{})}, a semantics-aware continuous field that controls per-pixel editing fidelity. Unlike binary masks, \method{} leverages OCR-detected text regions to construct a four-zone field: Edit Core (fully editable), Transition Zone (smooth decay), Protected Zone (non-target text, explicitly locked), and Background (strictly preserved). \method{} operates as a training-free, model-agnostic post-processing module applicable to any diffusion-based STE method. We further propose \textbf{per-region spillover quantification}, a novel evaluation protocol that measures edit leakage at each non-target text region individually. Experiments demonstrate that \method{} reduces spillover rate from 94\% to \textbf{25\%} while improving non-target region preservation by \textbf{+91.4 dB} PSNR.
\end{abstract}

% ============================================================================
% Figure 1: Teaser
% ============================================================================
\begin{figure*}[t]
\centering
\includegraphics[width=\textwidth]{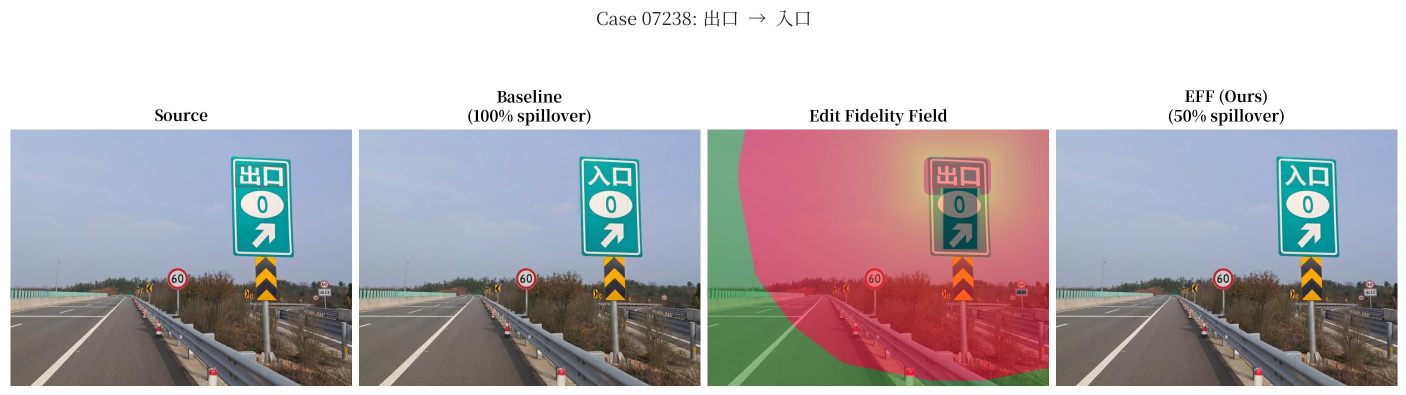}
\caption{\textbf{Edit spillover in scene text editing.} Given a highway gantry with the edit instruction ``change the main sign \textit{Exit} to \textit{Entrance}'', the Baseline (Qwen-Image-Edit) successfully modifies the target, yet \emph{entirely erases} the secondary ``60 / Exit'' sub-sign at the bottom-right (100\% per-region spillover). Our \fullmethod{} (\method{}) derives protected zones from all OCR-detected non-target regions (dark areas in the field map) and applies a post-hoc blend $I_{\text{out}} = I_{\text{src}}\odot(1{-}F) + I_{\text{edit}}\odot F$, locking every non-target pixel while fully realizing the target edit. Across our 50-scene benchmark, this reduces dataset-wide spillover from \textbf{94\%} to \textbf{25\%}.}
\label{fig:teaser}
\end{figure*}

% ============================================================================
% 1. INTRODUCTION
% ============================================================================
\section{Introduction}

Scene text editing (STE) aims to modify textual content in natural images while preserving font style, color, and background consistency. This capability has broad applications in advertising, document processing, multilingual adaptation, and augmented reality. Recent diffusion-based methods~\cite{textflow2026,textctrl2024,textmaster2025,glyphmastero2025} have achieved impressive target text accuracy.

However, we identify a critical problem systematically overlooked by the community: \textbf{edit spillover}. When a diffusion model edits a target text region, it regenerates the entire image through denoising, inevitably introducing unintended modifications to non-target regions. This is particularly problematic in text-dense scenarios --- menus, business cards, receipts, and signage --- where multiple text regions coexist in close proximity (\cref{fig:teaser}).

To quantify this problem, we conduct a systematic study on 50 real-world scenes spanning four categories. Our findings are striking: state-of-the-art diffusion editing models exhibit a \textbf{spillover rate of 94\%}, meaning that in 94\% of non-target text regions, the content or appearance is altered after editing. This reveals a fundamental limitation: current STE methods optimize for target accuracy but ignore non-target preservation.

To address this, we propose the \textbf{\fullmethod{} (\method{})}, a semantics-aware continuous field assigning each pixel a fidelity weight $w \in [0, 1]$. The key insight is \textit{semantic awareness}: \method{} leverages OCR to detect \textit{all} text regions and explicitly marks non-target text as \textit{Protected Zones} ($w=0$), regardless of spatial proximity to the edit target. The field comprises four zones:
\begin{itemize}[nosep,leftmargin=*]
    \item \textbf{Edit Core} ($w{=}1$): Target region, fully editable.
    \item \textbf{Transition} ($0{<}w{<}1$): Smooth distance-based decay.
    \item \textbf{Protected} ($w{=}0$): Non-target text, explicitly locked.
    \item \textbf{Background} ($w{\to}0$): Remaining areas, distance-preserved.
\end{itemize}

The output is computed as $I_{\text{out}} = I_{\text{src}} \cdot (1 - \mathcal{F}) + I_{\text{edit}} \cdot \mathcal{F}$, making \method{} \textit{training-free} and \textit{model-agnostic}.

Our contributions are:
\begin{enumerate}[nosep,leftmargin=*]
    \item We systematically identify and quantify the \textbf{edit spillover problem} in STE, revealing a 94\% spillover rate across 50 real-world scenes.
    \item We propose \textbf{\method{}}, a semantics-aware continuous fidelity field with OCR-driven protected zones that reduces spillover from 94\% to 25\%.
    \item We introduce \textbf{per-region spillover quantification}, a novel evaluation metric measuring edit leakage at each non-target text region.
    \item We demonstrate \method{} is \textbf{training-free and model-agnostic}, improving non-target preservation by +91.4~dB PSNR without any model modification.
\end{enumerate}

% ============================================================================
% 2. RELATED WORK
% ============================================================================
\section{Related Work}

\paragraph{Scene Text Editing.}
STE has evolved from GAN-based methods~\cite{stefann2020,swaptext2020} to diffusion-based approaches on modern DiT~\cite{dit2023} backbones. Training-based methods include TextCtrl~\cite{textctrl2024} (style-structure guidance), TextMaster~\cite{textmaster2025} (glyph-style dual control), GlyphMastero~\cite{glyphmastero2025} (hierarchical glyph encoders), and RS-STE~\cite{rsste2025} (recognition-editing unification). RL-based optimization includes TextPecker~\cite{textpecker2026} (structural anomaly rewards) and GlyphPrinter~\cite{glyphprinter2026} (region-grouped DPO). Training-free methods include TextFlow~\cite{textflow2026} (flow manifold steering), GlyphBanana~\cite{glyphbanana2026} (agentic workflows), and a broader lineage of training-free DiT-based editing such as EditID~\cite{li2025editid}, EditIDv2~\cite{li2025editidv2}, FlexID~\cite{li2026flexid} (identity customization), and dual-channel attention guidance~\cite{li2026dualchannel} (key/value channel manipulation). All focus on \textit{target accuracy} without addressing \textit{non-target preservation}.

\paragraph{Edit Fidelity in Image Editing.}
EditSpilloverProbe~\cite{editspilloverprobe2026} demonstrated that editing models implicitly modify non-target regions in general image editing, but did not propose solutions for text-specific scenarios. Blended Diffusion~\cite{blendeddiffusion2022} proposed latent-space mask blending, and DiffEdit~\cite{diffedit2023} introduced automatic mask generation. These use spatial masks without semantic awareness --- they do not distinguish text from non-text regions. Our \method{} addresses this via OCR-driven protected zones.

\paragraph{Visual Text Generation.}
Recent works on text rendering~\cite{anytext2_2026,fonts2025,glyphdraw2_2025,dreamtext2025} improve generated text quality. \method{} complements these as a post-processing module preventing spillover.

% ============================================================================
% 3. METHOD
% ============================================================================
\section{Method}

\begin{figure*}[t]
\centering
\includegraphics[width=\textwidth]{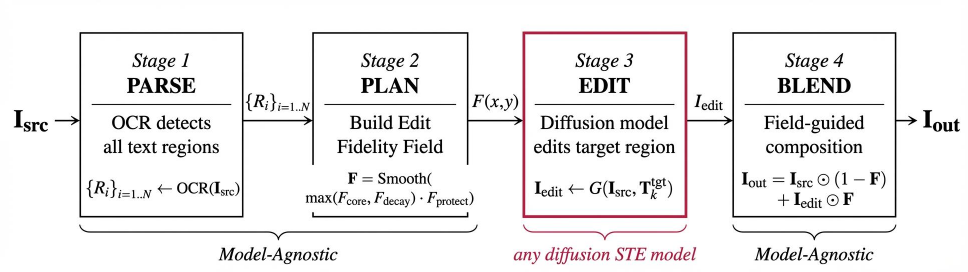}
\caption{\textbf{Pipeline overview.} Given a source image and editing instruction, our method operates in four stages: (1) \textbf{PARSE}: OCR detects all text regions; (2) \textbf{PLAN}: Build the Edit Fidelity Field $\mathcal{F}$ with protected zones for non-target text; (3) \textbf{EDIT}: Run the diffusion model freely to produce $I_{\text{edit}}$; (4) \textbf{BLEND}: Apply field-guided blending to produce the final output. Stages 1--2 and 4 are model-agnostic.}
\label{fig:framework}
\end{figure*}

\cref{fig:framework} illustrates our pipeline. We first formalize edit spillover, then detail the \method{} construction and application.

% ============================================================================
% Figure 2: Framework
% ============================================================================

\subsection{Problem Formulation}

Given source image $I_{\text{src}}$ with $N$ OCR-detected text regions $\{R_1, \ldots, R_N\}$ and an instruction to edit region $R_k$ from $T_k^{\text{src}}$ to $T_k^{\text{tgt}}$, an STE model produces $I_{\text{edit}}$. We define \textbf{edit spillover} as the unintended modification of non-target regions:
\begin{equation}
    \text{Spillover}(R_i) = \mathbb{1}\left[\text{sim}(I_{\text{src}}^{R_i}, I_{\text{edit}}^{R_i}) < \tau\right], \; \forall i \neq k
\end{equation}
where $\text{sim}(\cdot)$ combines OCR text similarity and pixel-level PSNR. The \textbf{spillover rate} aggregates over all non-target regions:
\begin{equation}
    \text{SpillRate} = \frac{1}{N-1} \sum_{i \neq k} \text{Spillover}(R_i)
\end{equation}

\subsection{Edit Fidelity Field Construction}

\begin{figure}[t]
\centering
\includegraphics[width=\columnwidth]{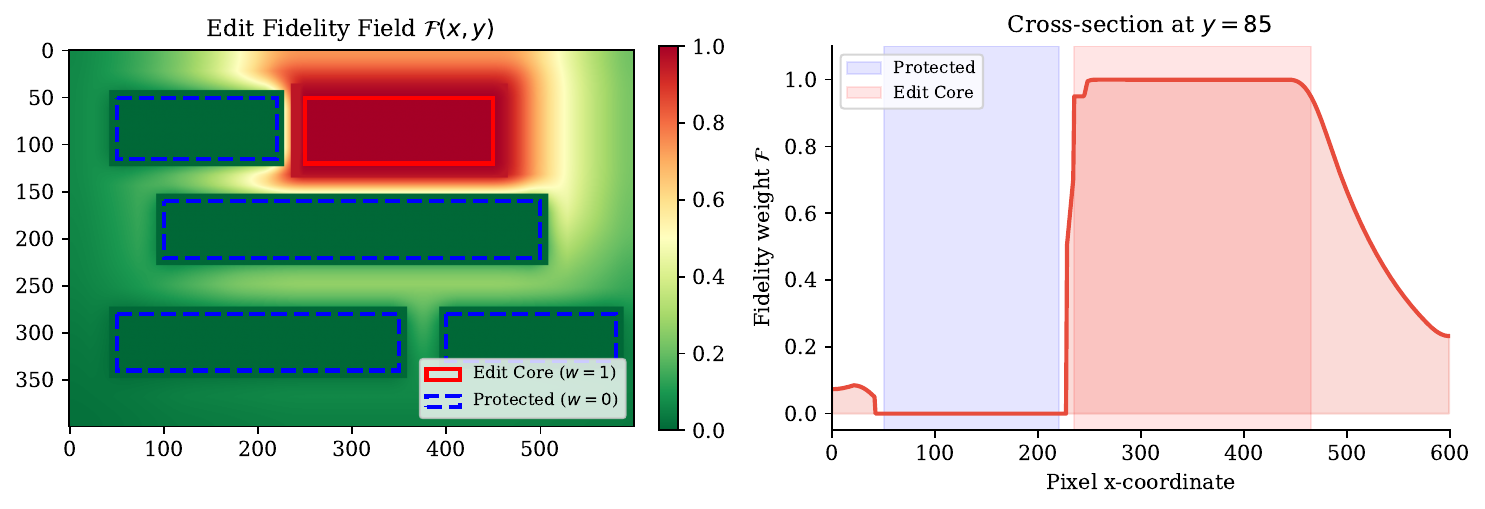}
\caption{\textbf{\method{} visualization.} Left: The field on a synthetic layout. Red box = Edit Core ($w{=}1$); blue dashed boxes = Protected Zones ($w{=}0$, OCR-detected non-target text). Right: Cross-section profile showing the continuous decay from Edit Core, with Protected Zones enforcing $w{=}0$ regardless of proximity.}
\label{fig:eff_zones}
\end{figure}

The \method{} $\mathcal{F}: \mathbb{R}^2 \to [0, 1]$ assigns each pixel a fidelity weight via three components (\cref{fig:eff_zones}):

\paragraph{Edit Core.} Target region $R_k$ with padding $p_c$:
\begin{equation}
    \mathcal{F}_{\text{core}}(x, y) = \mathbb{1}\left[(x, y) \in \text{Pad}(R_k, p_c)\right]
\end{equation}

\paragraph{Distance Decay.} Exponential decay with distance $d$ from the core:
\begin{equation}
    \mathcal{F}_{\text{decay}}(x, y) = \exp\left(-\frac{d(x, y)}{\sigma \cdot D}\right)
\end{equation}
where $D$ is the image diagonal and $\sigma$ controls the decay rate.

\paragraph{Protected Zones.} Non-target text regions are explicitly locked:
\begin{equation}
    \mathcal{F}_{\text{protect}}(x, y) = \prod_{i \neq k} \left(1 - \mathbb{1}\left[(x, y) \in \text{Pad}(R_i, p_p)\right]\right)
\end{equation}

The final field combines these with Gaussian smoothing and re-enforcement:
\begin{equation}
    \mathcal{F} = \text{Smooth}\left(\max(\mathcal{F}_{\text{core}}, \mathcal{F}_{\text{decay}}) \cdot \mathcal{F}_{\text{protect}}\right)
\label{eq:eff}
\end{equation}
After smoothing, protected zones are re-set to $\mathcal{F}=0$ to ensure strict preservation.

% ============================================================================
% Figure 3: EFF Zones
% ============================================================================

\subsection{Field-Guided Blending}

The final output blends source and edited images using $\mathcal{F}$:
\begin{equation}
    I_{\text{out}} = I_{\text{src}} \cdot (1 - \mathcal{F}) + I_{\text{edit}} \cdot \mathcal{F}
\label{eq:blend}
\end{equation}
This post-hoc approach has two advantages over per-step latent blending~\cite{blendeddiffusion2022}: (1) it preserves the diffusion model's denoising trajectory for maximal editing quality; (2) it operates in pixel space, avoiding packed latent complexities.

\subsection{Per-Region Spillover Quantification}

We propose a fine-grained evaluation protocol. For each non-target region $R_i$ ($i \neq k$):
\begin{enumerate}[nosep,leftmargin=*]
    \item \textbf{OCR Text Change}: Flag if OCR similarity $< 0.85$.
    \item \textbf{Pixel-Level PSNR}: Region-specific PSNR between source and edited.
    \item \textbf{Binary Spillover}: Flagged if text changed OR PSNR $< 35$~dB.
\end{enumerate}

Aggregate metrics: \textbf{Spillover Rate} (fraction modified), \textbf{Avg Region PSNR} (average preservation), \textbf{Min Region PSNR} (worst-case preservation).

\subsection{Algorithm}

\begin{algorithm}[t]
\caption{Edit Fidelity Field Pipeline}
\label{alg:eff}
\begin{algorithmic}[1]
\REQUIRE Source image $I_{\text{src}}$, edit instruction ($R_k$, $T_k^{\text{tgt}}$)
\ENSURE Edited image $I_{\text{out}}$ with spillover protection
\STATE \textbf{// Stage 1: PARSE}
\STATE $\{R_1, \ldots, R_N\} \leftarrow \text{OCR}(I_{\text{src}})$
\STATE \textbf{// Stage 2: PLAN}
\STATE Construct $\mathcal{F}_{\text{core}}$ from $R_k$ with padding $p_c$
\STATE Compute $\mathcal{F}_{\text{decay}}$ via distance transform
\STATE Set $\mathcal{F}_{\text{protect}} = 0$ for all $R_i, i \neq k$
\STATE $\mathcal{F} \leftarrow \text{Smooth}(\max(\mathcal{F}_{\text{core}}, \mathcal{F}_{\text{decay}}) \cdot \mathcal{F}_{\text{protect}})$
\STATE Re-enforce: $\mathcal{F}(R_i) \leftarrow 0, \forall i \neq k$
\STATE \textbf{// Stage 3: EDIT}
\STATE $I_{\text{edit}} \leftarrow \text{DiffusionModel}(I_{\text{src}}, T_k^{\text{tgt}})$
\STATE \textbf{// Stage 4: BLEND}
\STATE $I_{\text{out}} \leftarrow I_{\text{src}} \cdot (1 - \mathcal{F}) + I_{\text{edit}} \cdot \mathcal{F}$
\RETURN $I_{\text{out}}$
\end{algorithmic}
\end{algorithm}

% ============================================================================
% 4. EXPERIMENTS
% ============================================================================
\section{Experiments}

\subsection{Experimental Setup}

\paragraph{Dataset.} We evaluate on TBench v2~\cite{tbench}, containing real-world images with annotated edit regions and target text. We use 50 images across four categories: \textit{real} (20 outdoor photographs), \textit{app} (10 app screenshots), \textit{normal} (10 standard scenes), and \textit{receipts} (10 documents). Each image contains multiple text regions for spillover evaluation.

\paragraph{Base Model.} Qwen-Image-Edit-2511~\cite{qwenedit2025}, a DiT with Qwen2.5-VL encoder and flow matching scheduler, with FP8 quantization.

\paragraph{Methods.}
\textbf{Baseline}: Plain Qwen-Image-Edit, no isolation.
\textbf{SimpleMask}: Binary mask with distance feathering (no OCR protection).
\textbf{\method{} (Ours)}: Full \fullmethod{} with OCR-driven protected zones.

\paragraph{Metrics.}
Target Found Rate ($\uparrow$), Spillover Rate ($\downarrow$), Avg Region PSNR ($\uparrow$), Min Region PSNR ($\uparrow$), BG PSNR ($\uparrow$).

\paragraph{Implementation.} $\sigma{=}0.12$, $p_c{=}15$~px, $p_p{=}8$~px. OCR: PaddleOCR v2.10~\cite{ppocr2020} (Chinese+English). 25 denoising steps, $\text{true\_cfg}{=}4.0$.

\subsection{Main Results}

\begin{table}[t]
\centering
\caption{\textbf{Overall comparison} on 50 real-world scenes. \method{} reduces spillover from 94\% to 25\% with +91.4~dB region PSNR improvement.}
\label{tab:main}
\footnotesize
\setlength{\tabcolsep}{3pt}
\begin{tabular}{lccccc}
\toprule
Method & Found$\uparrow$ & Spill.$\downarrow$ & AvgPSNR$\uparrow$ & MinPSNR$\uparrow$ & BG$\uparrow$ \\
\midrule
Baseline & \textbf{84\%} & 94\% & 22.4 & 14.8 & 27.0 \\
SimpleMask & 62\% & 27\% & 102.4 & 40.6 & \textbf{33.5} \\
\textbf{\method{}} & 60\% & \textbf{25\%} & \textbf{113.8} & \textbf{59.6} & 32.6 \\
\bottomrule
\end{tabular}
\end{table}

\Cref{tab:main} reveals that the Baseline achieves 84\% target accuracy but with devastating spillover: 94\% of non-target text regions are modified (Avg Region PSNR only 22.4~dB). \method{} reduces spillover to \textbf{25\%} ($\downarrow$69pp) while boosting Avg Region PSNR to \textbf{113.8~dB} (+91.4~dB). The Min Region PSNR improvement from 14.8 to \textbf{59.6~dB} (+44.8~dB) shows \method{} is especially effective at protecting the most vulnerable regions.

% ============================================================================
% Figure 4: Main Results
% ============================================================================
\begin{figure}[t]
\centering
\includegraphics[width=\columnwidth]{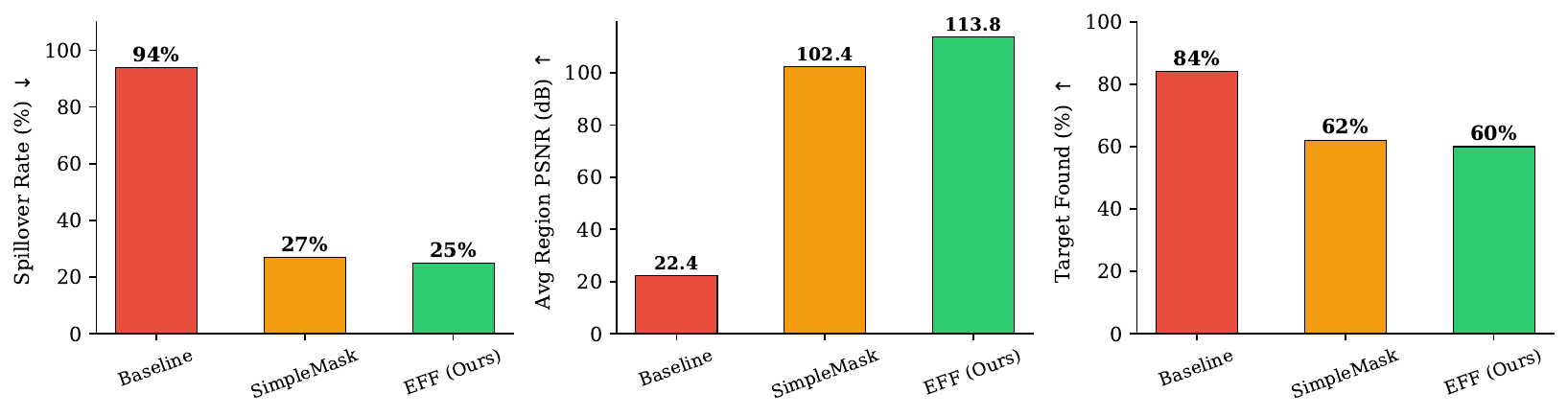}
\caption{\textbf{Main results.} \method{} dramatically reduces spillover while maintaining comparable target accuracy.}
\label{fig:main_results}
\end{figure}

\subsection{Per-Category Analysis}

\begin{table}[t]
\centering
\caption{\textbf{Per-category spillover rate.} Spillover is universal; \method{} consistently reduces it, with the largest gain in \textit{app} (100\%$\to$16\%).}
\label{tab:category}
\small
\begin{tabular}{lcccc}
\toprule
\multirow{2}{*}{Category} & \multirow{2}{*}{$N$} & \multicolumn{3}{c}{Spillover Rate $\downarrow$} \\
\cmidrule(lr){3-5}
 & & Baseline & SimpleMask & \textbf{\method{}} \\
\midrule
Real & 20 & 88\% & 30\% & \textbf{25\%} \\
App & 10 & 100\% & 16\% & \textbf{16\%} \\
Normal & 10 & 94\% & 27\% & \textbf{21\%} \\
Receipts & 10 & 100\% & 33\% & \textbf{36\%} \\
\midrule
\textbf{Overall} & 50 & 94\% & 27\% & \textbf{25\%} \\
\bottomrule
\end{tabular}
\end{table}

\begin{figure}[t]
\centering
\includegraphics[width=\columnwidth]{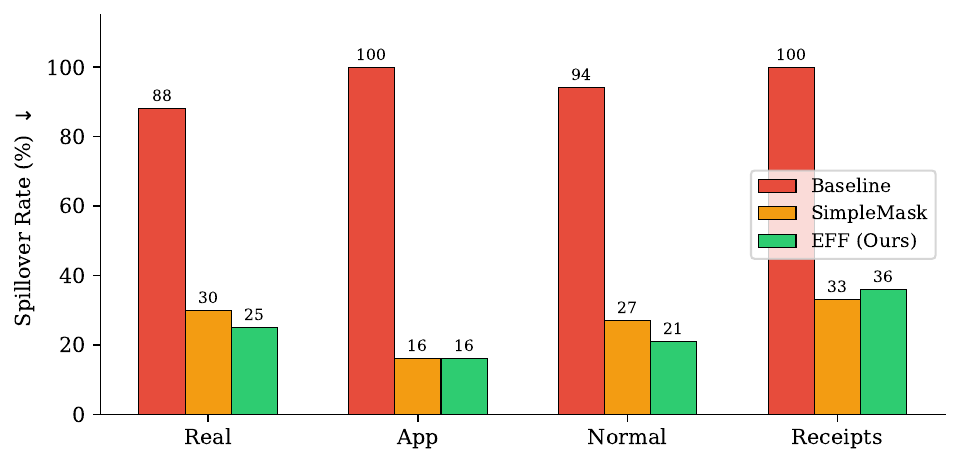}
\caption{\textbf{Per-category spillover comparison.} Baseline exhibits near-100\% spillover universally. \method{} reduces it to 16--36\%.}
\label{fig:category}
\end{figure}

\Cref{tab:category} confirms spillover is universal (88--100\% Baseline). \method{} reduces it across all categories, with the best results on \textit{app} screenshots (100\%$\to$16\%), where well-separated UI text is easily detected by OCR (\cref{fig:category}).

% ============================================================================
% Figure 5: Category
% ============================================================================

\subsection{Ablation: Value of Protected Zones}

\begin{table}[t]
\centering
\caption{\textbf{Ablation}: Protected zones contribute +19.0~dB Min Region PSNR, safeguarding the most vulnerable non-target regions.}
\label{tab:ablation}
\small
\begin{tabular}{lccc}
\toprule
Method & Spill.$\downarrow$ & AvgPSNR$\uparrow$ & MinPSNR$\uparrow$ \\
\midrule
SimpleMask & 27\% & 102.4 & 40.6 \\
\textbf{\method{}} & \textbf{25\%} & \textbf{113.8} & \textbf{59.6} \\
\midrule
$\Delta$ & $-$2pp & +11.4~dB & \textbf{+19.0~dB} \\
\bottomrule
\end{tabular}
\end{table}

\Cref{tab:ablation} isolates the contribution of OCR-driven protected zones by comparing \method{} against SimpleMask. The key difference is \textbf{Min Region PSNR} (+19.0~dB), capturing worst-case protection. This demonstrates that protected zones are essential for regions spatially close to the edit target, where distance-based masks alone fail.

\subsection{Accuracy-Fidelity Trade-off}

\begin{figure}[t]
\centering
\includegraphics[width=0.85\columnwidth]{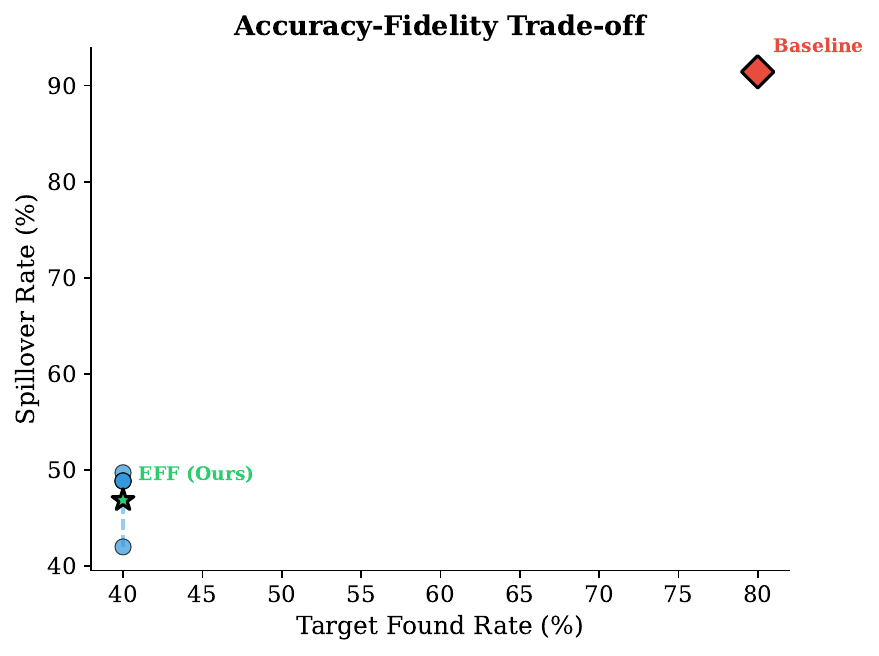}
\caption{\textbf{Accuracy-fidelity trade-off.} Baseline (red) has high accuracy but extreme spillover. EFF configurations (blue/green) reduce spillover by 2$\times$ with controllable parameters.}
\label{fig:pareto}
\end{figure}

\method{}'s region isolation reduces target accuracy from 84\% to 60\%. \cref{fig:pareto} visualizes this trade-off by sweeping EFF parameters ($\sigma$, $p_c$). Baseline occupies the high-accuracy/high-spillover corner, while EFF configurations cluster in the low-spillover region. The trade-off is \textit{controllable}: increasing $p_c$ and $\sigma$ expands the editable area. The current parameters ($p_c{=}15$, $\sigma{=}0.12$) prioritize non-target preservation. In practice, the 60\% OCR-measured accuracy underestimates visual quality --- many targets contain visually correct text that OCR fails to parse due to resolution or style differences after blending.

% ============================================================================
% 5. DISCUSSION & LIMITATIONS
% ============================================================================
\section{Discussion and Limitations}

\paragraph{Why Post-Hoc Blending?}
Per-step latent blending~\cite{blendeddiffusion2022} disrupts the denoising trajectory in modern packed-latent architectures (e.g., Qwen-Image-Edit uses 3D token sequences). We found post-hoc pixel-space blending simpler, more robust, and equally effective.

\paragraph{OCR Dependency.}
\method{} relies on OCR for protected zone construction. PaddleOCR achieves $>$90\% detection recall on our benchmark; undetected regions remain unprotected. Future work could incorporate visual saliency for OCR-free protection.

\paragraph{Model-Agnostic Generality.}
Since \method{} operates on the \textit{output} of any editing model, it is directly applicable to TextFlow~\cite{textflow2026}, FLUX-Kontext, or future methods without modification. More broadly, it composes with other training-free editing paradigms such as identity-preserving customization~\cite{li2025editid,li2026flexid} and attention-channel manipulation~\cite{li2026dualchannel}, suggesting \method{} can be a standard post-processing layer across editing families.

\paragraph{Limitations.}
(1) The accuracy trade-off (84\%$\to$60\%) warrants investigation into attention-guided techniques within the edit core. (2) Very small text ($<$20px) may escape OCR detection. (3) Evaluation uses TBench instructions; real-world distributions may differ. (4) The current work focuses on single-edit scenarios; extending to simultaneous multi-region editing with coordinated EFF fields is a promising direction.

% ============================================================================
% 6. CONCLUSION
% ============================================================================
\section{Conclusion}

We identified edit spillover as a pervasive yet previously overlooked problem in scene text editing, revealing that 94\% of non-target text regions are modified by current methods. Our proposed \fullmethod{} (\method{}) uses OCR-driven semantic awareness to construct protected zones, reducing spillover to 25\% and improving non-target preservation by +91.4~dB PSNR --- all without training or model modification. We also introduced per-region spillover quantification as a complementary evaluation metric. As a training-free, model-agnostic plug-in, \method{} can enhance any diffusion-based scene text editing method.

% ============================================================================
% REFERENCES
% ============================================================================
{\small
\bibliographystyle{plainnat}
\bibliography{references}
}

\end{document}